# Performance Evaluation of a ROS2 Based Automated Driving System

Jorin Kouril[1], Bernd Schäufele[1], Ilja Radusch[2] and Bettina Schnor[3]

[1]*Fraunhofer Institute for Open Communication Systems (FOKUS), Berlin, Germany*
[2]*Daimler Center for Automotive Information Technology Innovations (DCAITI), Berlin, Germany*
[3]*Department of Computer Science, University of Potsdam, Potsdam, Germany*

Abstract: Automated driving is currently a prominent area of scientific work. In the future, highly automated driving and new Advanced Driver Assistance Systems will become reality. While Advanced Driver Assistance Systems and automated driving functions for certain domains are already commercially available, ubiquitous automated driving in complex scenarios remains a subject of ongoing research. Contrarily to single-purpose Electronic Control Units, the software for automated driving is often executed on high performance PCs. The Robot Operating System 2 (ROS2) is commonly used to connect components in an automated driving system. Due to the time critical nature of automated driving systems, the performance of the framework is especially important. In this paper, a thorough performance evaluation of ROS2 is conducted, both in terms of timeliness and error rate. The results show that ROS2 is a suitable framework for automated driving systems.

## 1 INTRODUCTION

Automated driving is a trending area of research, with a lot of effort from both academia and economy. Modern vehicles are equipped with many Advanced Driver Assistance Systems (ADAS) and even automated driving functions, such as highway pilots (Binder et al., 2016). Despite the existence of these systems, automated driving higher than level 3 (SAE, 2021) is still a challenge, especially in complex and urban environments.

In most commercial cars, there are many different Electronic Control Units (ECU), each for a specific purpose. Vehicle bus systems, most prominently CAN and FlexRay (Reif, 2011), allow these systems to communicate with each other. Research in automated driving shows that the tasks for this challenge are more complex and have high computational requirements. For example, the evaluation of sensor values to detect objects, is often performed with neural networks (Spielberg et al., 2019). Also, some of the tasks regarding automated driving are linked, e.g., object detection and localization.

For the development of automated driving systems (ADS), the Robot Operating System (ROS) is widely used (Reke et al., 2020), especially the Robot Operating System 2 (ROS2). The tasks regarding automated driving can be more efficiently performed on one or several central computation units, e.g., high performance PCs. As the realization of automated driving consists of several subtasks, a modular software architecture is suitable. An automated vehicle must sense and detect other objects, it must localize itself, and it must plan and control a trajectory. ROS2 facilitates simple communication between modules through a publish and subscribe pattern. As it was designed for the development of robots, it also provides a rich ecosystem of useful libraries for automated driving, such as probabilistic filters and planning algorithms.

ROS2 abstracts middleware communication across several levels in a high-level API. The foundation for message exchange is a data distribution service (DDS), defined by the standard of the same name (Object Management Group, 2015). The connection between the DDS and ROS2 is abstracted using the ROS middleware interface (rmw). The core functionality of ROS2 is implemented in the ROS client library (rcl), which is based on the rmw. Applications are normally implemented in language-specific wrappers of the rcl.

The control of an automated vehicle imposes severe temporal and reliability requirements. The detection of obstacles and the planning of a path algorithm must be completed within a certain time frame. Furthermore, information must not be lost. Having a modular architecture with a distributed framework, such as ROS2, demands that the framework itself is performing efficiently. The high amount data that is necessary to be processed for automated driving, e.g., LIDAR point clouds and camera streams, makes this task even more challenging.

Therefore, in this paper, a performance evaluation of the ROS2 framework in an automated vehicle is presented. Particularly, the suitability of different middleware implementations for vehicular applications is investigated. These implementations are compared in terms of latency and error susceptibility. In this context, latency refers to the time elapsed from message transmission to reception. Besides, the error rate is quantified as packet loss. The scenarios for the analysis vary in terms of the number of components in the graph and the size and frequency of individual data packets. All evaluations are performed on an actual on-board PC in an automated vehicle.

The paper is structured as follows: In section 2, an overview of the related work is given. Section 3 explains the ADS used for the evaluation in detail. The implementation of the software for evaluation is shown in section 4. Consequently, in section 5 the results are presented and discussed, before an overview and outlook is given in section 6.

## 2 RELATED WORK

Several architectures for automated driving based on ROS or ROS2 exist. Some of the most prominent ones are Autoware.auto (The Autoware Foundation, 2023) and Apollo (Baidu Apollo consortium, 2023). These systems show that ROS2 is a suitable framework for developing an ADS. However, due to their complexity, for research purposes, more lightweight approaches can lead to faster results and better performance. An analysis of the performance of Autoware.auto yields good results, but this is not generalized to ROS2 (Li et al., 2022).

In another publication, an alternative architecture for a ROS2 based automated vehicle is presented (Reke et al., 2020). The system is described in detail and a performance evaluation is presented. This work indicates that ROS2 is suitable for real time operations. However, the Data Distribution Services (DDS) is not exchanged for analysis, and packet loss is not examined, either.

An assessment of the performance of ROS2 took place very early in the development stage (Maruyama et al., 2016). Here, a comparative analysis is conducted between ROS1 and ROS2 to assess the potential positive impact of the novel concepts introduced in ROS2. At that point of time, ROS2 does not exhibit superior performance compared to ROS1. However, a notable improvement can be observed, particularly regarding the equal distribution of latencies across all subscribers.

A different study investigates the real-time capabilities of ROS2 (Gutiérrez et al., 2018). The evaluation focuses on the ability of ROS2 to achieve soft real-time capabilities, indicating its potential for applications with timing constraints. The evaluation methodology primarily considers one-to-one communication, while more complex many-to-many scenarios are substituted with artificially generated workloads external to the ROS2 applications. This approach allows for an assessment of the performance of ROS2 in a controlled environment.

In a more recent work, the performance of the three official DDS implementations (FastDDS, CycloneDDS, and RTI Connext) is compared, varying sending frequencies, packet sizes, and participants (Kronauer et al., 2021). Consistent with (Maruyama et al., 2016), it is observed that latency exhibits a sharp increase beyond the UDP fragment size of 64kB. Furthermore, the authors conclude that DDS is the primary contributor to latency.

## 3 AUTOMATED DRIVING SYSTEM

For automated driving, Fraunhofer FOKUS uses a hybrid Mercedes E-Class, which is able to plan and drive paths in an automated way. The vehicle is used to developed different ADS, such as automated valet parking (Schäufele et al., 2017). It is equipped with communication hardware for cooperative maneuvers as well (Schaeufele et al., 2017; Eiermann et al., 2020).

Due to its complexity, the overall system is divided in subsystems. As a result of the modular architecture, ROS2 was selected, because it allows for simple communication between components through a publish and subscribe mechanism. Besides, ROS2 (Macenski et al., 2022) offers many robotic libraries that can be applied for an ADS. ROS2 is used to implement the components of the architecture.

The design of the system follows the pattern of Sense, Plan, Act (During and Lemmer, 2016). First, a representation of the environment of the vehicle is created with sensing. The sensors are evaluated in the Perception Unit (PU), which is an on-board PC with high performance graphics hardware. In the planning stage, the environment model and other constraints, such as vehicle parameters, are used for the calculation of a drivable trajectory for the vehicle. In acting, the planned trajectory is controlled and executed.

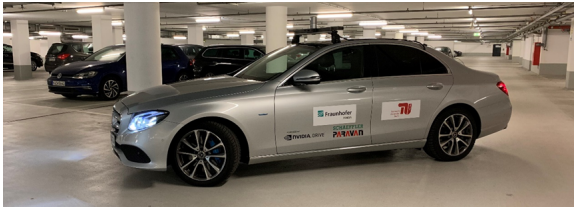

Figure 1: The automated vehicle of Fraunhofer FOKUS.

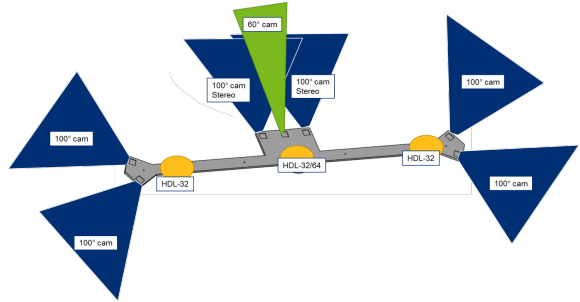

Figure 2: Overview of the sensor rig.

The perception system of the automated vehicle, called 3D Vision, allows full understanding of the surroundings. The test vehicle can be seen in Figure 1. For the 3D Vision, the car is equipped with a sensor rig that can hold various sensors. A schematic overview of the sensor rig is shown in Figure 2.

The sensor setup consists of three LIDAR scanners, which create a 3D point cloud of the vehicle surroundings. For a full view in camera images, seven cameras are mounted on the sensor rig, one camera with a 60 degrees aperture to the front, four cameras with 100 degrees aperture on the corners, and additionally two front cameras with 100 degrees aperture, which provide stereo images.

The sensor evaluation is performed with neural networks at the PU. In an early fusion, LIDAR points are projected onto the 2D camera images, which are processed with Convolutional Neural Networks (CNN). Due to the projection, the 3D coordinates of the object detections from the camera images can be determined. Figure 3 shows the results from the image processing. The network detects various traffic objects, such as cars, scooters, and traffic signs. For lane detection, a novel early fusion approach is implemented (Wulff et al., 2018).

For LIDAR perception, the points are grouped in bins and various features are calculated for each bin with neural networks, such as Pointpillars (Lang et al., 2019) and SECOND (Yan et al., 2018). These networks can process data more efficiently compared to raw point clouds. This efficiency stems from the utilization of an internal representations in the form of bins, enabling faster processing while still yielding valuable outcomes.

The perception results are shown in Figure 4. The top left, bottom left, and bottom center show the 2D bounding boxes in camera images. The top center shows an internal representation of the LIDAR processing, in which each LIDAR point is assigned to a specific bin. The 3D bounding boxes are shown in the top right. The hardware setup can be seen in Figure 5 with the PU rack in the center and devices for sensor connection and vehicular communication.

The objects derived from camera and LIDAR are collected in an environment model. It takes care of tracking objects, i.e., assigning a unique identifier over consecutive time frames. Thereby, the environment model fuses the object detections from the different sensors to a single internal object representation. The output of the environment model is passed to the planning and acting stages.

The path planning builds upon the environment model and an existing route, which defines the vehicle's path at the road segment level, determining the segments to traverse and the turns to take at intersections. The route is map-based, but during path planning, it is enriched with real-time information from the perception and refined at lane level. The resulting path includes lane changes and avoids obstacles.

To further refine the path, a drivable trajectory is generated. This trajectory defines the desired position and time of the vehicle using a 2D spline, which is transmitted to the control system. The spline is continuous and adheres to vehicle constraints. It considers detected dynamic objects, such as vehicles and pedestrians, to ensure collision avoidance. Additionally, the trajectory is optimized to achieve efficient and comfortable driving.

For control, the vehicle is equipped with the Schaeffler Paravan Drive-by-Wire system (Unseld, 2020), which allows to actuate the steering wheel, and the throttle and brake controls. The control loop to follow the calculated spline is based on the pure pursuit algorithm (Samuel et al., 2016). The required steering wheel angles and forces are applied by the Drive-by-Wire system.

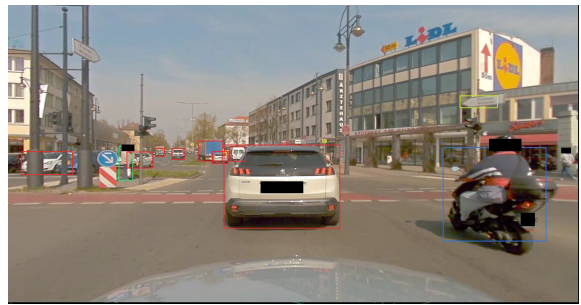

Figure 3: Results from camera perception.

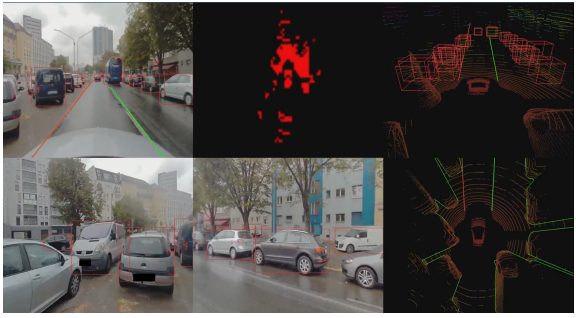

Figure 4: Results from LIDAR perception.

## 4 EVALUATION SYSTEM

To be able to correctly assess the performance of ROS2, the entire software stack must be taken into account, and influencing factors must be considered as isolated as possible. The aim of the measurements is to evaluate the performance of ROS2 regarding the requirements in an automated driving system. Two aspects are particularly essential for this. Is ROS2 fast enough to function as the backbone of interprocess communication in a real-time system? This condition can be measured very well with the latency, which for this work is defined as the elapsed time between sending the message and receiving the message in the user application. Another aspect to consider is the error rate in the system. Reliable delivery of messages is essential for an automated system, as the loss of information can lead to potentially dangerous wrong decisions. To estimate this metric, the occurring packet loss is measured as a percentage. Only local measurements are carried out for the measurement scenario, which corresponds to the current setup of the test vehicles used. The participants, nodes, and topics are also predefined; fluctuating behavior is not evaluated.

The full functional scope of ROS2 is realized via four different abstraction levels. Applications are written with the help of client libraries. These map the API in a specific programming language, officially supported here are C++ (rclcpp) and Python (rclpy). Most of the functionality is implemented in C and available as ROS client library. Communication with the specific DDS implementation, which manages the sending of messages and the discovery of other participants, is handled by the ROS middleware interface (rmw). Figure 6 summarizes the internal structure of ROS2. Each of the layers influences the overall performance.

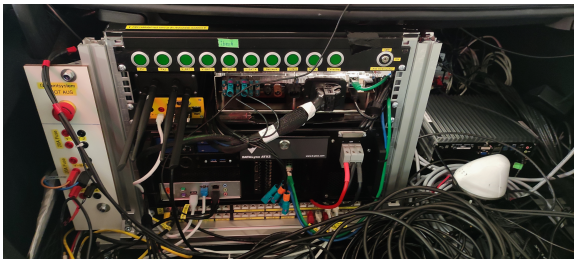

Figure 5: Hardware setup of the vehicle.

The first elementary influencing factor that comes into play is the DDS. Each DDS is used in its standard configuration to ensure basic comparability. The ROS2 stack with the rmw and rcl layer then follows, based on the DDS. The same version is used for each measurement to rule out possible deviations due to changes to these layers. rclcpp is used to implement the user applications for the measurement. In addition to the hardware and software stack, the way in which the system is used is also important. Parameters that have a major influence on the possible performance here are the data size per message, the number of messages sent per time unit, the number of nodes in the entire network, the number of topics used, and the number of publishers and subscribers per topic.

These parameters in particular are of great interest for the measurements, as they demand the central aspects of the DDS and ROS2 implementation with regard to their efficiency. Only the publish / subscribe pattern is considered for the measurements, as both services and actions are based on this methodology. Furthermore, the following assumptions are made for the measurements: A node is either a publisher or a subscriber, never both at the same time; each measurement is performed with one DDS; this is not exchanged during a measurement; and all nodes involved in the measurement are started beforehand;

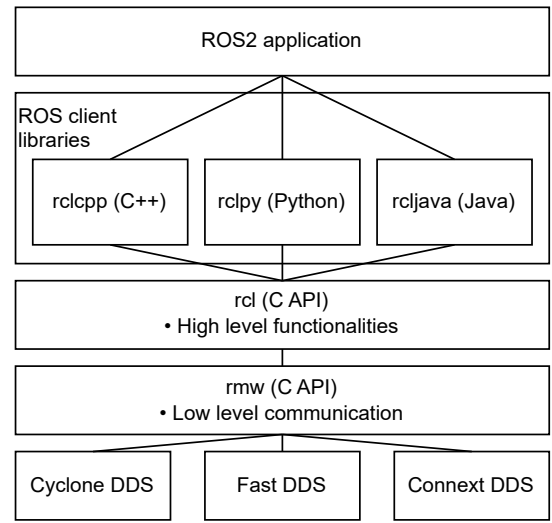

Figure 6: ROS2 internal architecture (Open Robotics, 2022a).

there are no late-joining components.

In a first iteration, the different DDS implementations are compared with each other. For this purpose, the performance test framework from Apex.AI (Pemmaiah et al., 2022) is used. Each DDS is tested in different scenarios with varying data sizes and numbers of participants. Based on the results, a selected DDS is then evaluated in detail to gain a deeper understanding of the performance of ROS2. For this purpose, a much larger number of factors are permuted and analyzed using tracing (Bédard et al., 2022) to track the path of the message through the software stack in order to precisely localize possible performance losses.

ROS2 works on the basis of workspaces (Open Robotics, 2022b). A workspace comprises a collection of ROS2 packages, i.e., ROS2-based software projects, which are built with the help of the ROS2-specific build tool Colcon (Open Robotics, 2022c). These workspaces can in turn build on each other so that the hierarchically higher workspace has access to all packages of the underlying workspace. This enables a structured and clean setup of all necessary packages without having to install packages that are not required for the specific measurement.

As first step of the implementation, the lowest common denominator of required packages is installed in a workspace. Here, this is the ROS2 library itself with the basic functionalities in the Rolling version. This workspace is built using the build tools, and all external dependencies are installed.

For the initial comparison testing a new workspace is created, and the Apex.AI Performance Test Project (Pemmaiah et al., 2022) is built. With the help of a Python script, a bash script is created from the benchmark configuration for the parameters. As part of the benchmark, the workspace is rebuilt once for each DDS so that the performance test uses it accordingly. All subscribers are started for each configuration and each DDS; the executable provided by Apex.AI is called up configured accordingly for this purpose. After a short wait to ensure that all subscribers have been initialized, the publishers are also started; the Apex.AI executable is also used here. After the configured time, the processes end automatically, and all open log files are closed. The process is repeated for each of the configurations. Only the layers below the rmw layer are used here. This allows to consider only the influence of the specific DDS implementation and to exclude any influences from higher layers.

Table 1 lists the permutations of the parameters; each of the DDS implementations is tested once for each of the specified configurations over a runtime of 60 seconds. The aim of the configurations is to ob-tain a comprehensive overview of the performance in order to be able to compare the various DDSs as accurately possible. The first three configurations serve as a basis for this. The simple 1 - 1 communication reduces the possible interference to a minimum. The 1 - 31 communication is already more demanding, as the DDS must now distribute the message to 31 subscribers, which leads to a considerably larger required bandwidth, especially with larger data packets. 32 nodes in a network comes much closer to an application in the field of automated driving in terms of the number of participants and can therefore provide initial indications of performance under load. The frequency set for all tests is 10Hz, which corresponds to the frequency most commonly used by sensors and computing components in automated driving systems and therefore serves as a sensible clock rate for generating the load. The different data sizes represent different scenarios. Struct16 is the smallest message and can, for example, be equated with a message from a simple sensor in the vehicle, such as acceleration. Array64k represents more complex data, such as a trajectory or recognized objects. Pointcloud1m is the largest message and is used to represent LIDAR scans or camera streams. This message size usually forms the upper limit of the messages used in the automated vehicle in terms of size per message. Overall, the parameters thus cover a good range from the actual application and provide initial indications of the performance of the various DDS implementations.

Table 1: Parameters of the comparative benchmark.

| Parameters | Values | |
|---|---|---|
| Nodes | 2 | 32 |
| Publisher Nodes | 1 | 1 |
| Subscriber Nodes | 1 | 31 |
| Size | Struct16, Array64k, PointCloud1m | |
| Frequency | 10Hz | |

Several ROS2 packages are required for more detailed measurements of a selected DDS. ROS2 tracing is elementary here (Bédard et al., 2022), as well as a special version of the DDS, to provide the necessary insight at this level. With the help of the ROS2 tracing package (Bédard et al., 2022) the message can be traced through the entire stack (Bédard et al., 2023) to understand how the latency arises, and where message losses occur. It is demonstrated that the additional overhead caused by tracing is minimal and therefore does not distort the results (Bédard et al., 2022).

For measurements, a package is required to generate the load in the system according to the configuration. For this purpose, a simple but fully configurable node is implemented as a publisher and subscriber. When the process is started, the publisher receives the frequency in milliseconds at which the message is to be sent, the size of the message in bytes, the length of the measurement in seconds and the topic on which it is to send. Based on this information, it then starts a ROS2 timer to periodically publish a packet on the topic according to the frequency. The message sent consists of a header, containing a timestamp and an ID in the form of an integer number, and a byte array of the defined size, which is filled with random values. The timestamp is entered in the message immediately before the actual transmission.

Table 2: Parameters of the detailed benchmark.

| Parameters | Values | | | |
|---|---|---|---|---|
| Nodes | 2 | 8 | 32 | 64 |
| Publisher Nodes | 1 | 1, 4, 7 | 1, 16, 31 | 1, 32, 63 |
| Subscriber Nodes | 1 | 1, 4, 7 | 1, 16, 31 | 1, 32, 63 |
| Subscriber per node | 1 | 1, 7 | 1, 31 | 1, 63 |
| Size | 0B, 64kB, 512kB, 1Mb, 2Mb | | | |
| Frequency | 10Hz, 100Hz | | | |

The structure of the subscriber nodes is similar. They receive a list of topics for which corresponding subscriptions are created. A timestamp is also taken directly in the associated callback, which is then compared with the timestamp from the received message, and the difference is saved. The actual measurement of the latency then follows using the recorded traces. The measurement is started with a Python script, which parses the various configurations and instantiates a corresponding number of publishers and subscribers, starts the tracing, and activates the corresponding DDS using an environment variable. The aim of these benchmarks is to gain a more detailed, in-depth insight into the performance of ROS2. For this purpose, the parameter space is significantly enlarged to obtain a higher resolution of the test results. The aim is also to go beyond the current requirements to be able to assess the performance limits of ROS2.

Table 2 shows the possible configurations of the parameters for the detailed benchmark. Based on the number of nodes, three topologies are created for the publisher-subscriber-subscriber-per-node ratio, as shown in Figure 7. The first topology has exactly as many publishers as subscribers, each with their own topic and one subscriber per node. The second topology changes the ratio, with only one publisher serving all the remaining nodes as a subscriber; here, too, there is only one subscriber per node. Finally, this relationship is reversed, and a node with many subscribers is served by the remaining nodes as a publisher. Topology 1 makes it possible to evaluate the influence of the number of topics on the overall performance and thus to test the scalability of topics. Topology 2 makes it possible to check how efficiently the distribution of messages on a topic works and how great the influence of the number of subscribers per topic is on the overall performance. Topology 3 allows to check how well the executor of a single node scales under load with more callbacks and how many subscribers per node can be effectively implemented before the SingleThreadedExecutor is overloaded. In practical applications, a mixture of all three topologies can be found. Each of the topologies is measured for each combination of frequency and data size over a runtime of 60 seconds.

For the detailed benchmarks, additional data sizes were added to the three data packets from the previous benchmarks. First, 512kB was added as the middle value of the previous value range to achieve better coverage in this area. Secondly, in order to test the limits, twice the previous maximum was added again with the aim of demanding the maximum bandwidth. Measurements at 100Hz were also added to the previously used frequency of 10Hz in order to gain an insight into the extent to which there is still potential for improvement here.

All benchmarks are run on the PU with an Intel® Xeon® E5-2667 v4 CPU and 8x 32GB RDIMM DDR4-2400+ reg ECC. An Intel X550-T2 network card handles communication with the connected sensors. The system is running Ubuntu 20.04 LTS.

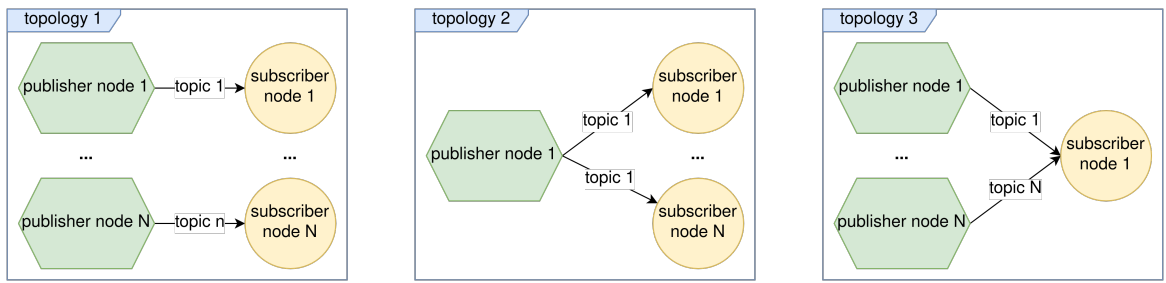

Figure 7: Overview of the different benchmark topologies.

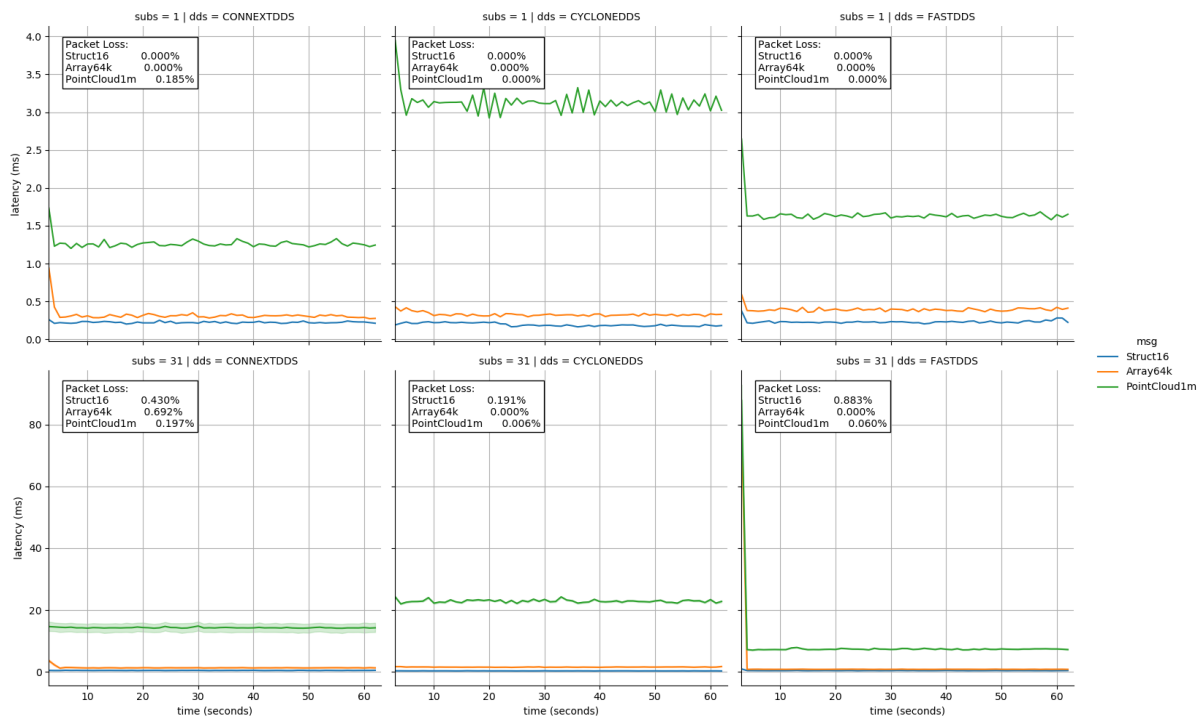

Figure 8: Comparison of latencies of DDS implementations over time.

## 5 EVALUATION RESULTS

A direct comparison of the three officially supported DDS implementations reveals differences but also similarities in performance behavior. Figure 8 shows the course of the latency in milliseconds over the measurement period of 60 seconds, with Struct16 in blue, Array64k in orange, and PointCloud1m in green as introduced in section 4. Each column represents one of the three DDS implementations. The top row shows the results for a publisher communicating with a subscriber. The bottom row shows the results for a publisher communicating with 31 subscribers. The first scenario here serves as a basic assessment, while the second scenario is more of an application scenario from the field of automated driving regarding the subscribers. The individual measurement lines represent the different packet sizes.

The measurement results are generally very good. For most scenarios, the latency remains below 2 ms, leaving a clear margin up to a frequency of 10 Hz. Only the latency for PointCloud1m is higher, which is particularly clear for the higher number of subscribers. In this case, the latency increases to at least 8ms (FastDDS) and on average to 15-20 ms. It is worth noting that the variance for each combination is below 1 ms, apart from ConnextDDS for 31 subscribers / PointCloud1m, where the variance is around 18 ms. This low variance indicates stable message transmission behavior, as does the almost constant latency over the course of the measurement.

The packet loss is also fairly limited. In the maximum case it is 0.88%, in most cases it is 0%. It is noticeable, however, that ConnextDDS is the only one with a packet loss of 0.18% in the 1-1 communication for the PointCloud1m messages, while all other 1-1 communication scenarios each have 0%. ConnextDDS also performs worse in the 1-31 scenario, losing a small number of packets for each message size. CycloneDDS and FastDDS predominantly lose packets for Struct16 in this case, which is presumably due to the fact that this packet is not fragmented. This means that the loss of a single UDP packet is not noticeable, whereas with fragmented messages there is a higher chance that at least one fragment will arrive and thus trigger a resend (Granados, 2017).

Overall, both latency and packet loss are satisfactory, even for more subscribers and larger data packets. This is illustrated again in Figure 9. The boxplot shows the latency per packet size for each of the three DDS. The box includes the upper and lower quartiles, the line within the box shows the median latency. The whiskers show the 1.5-fold quartile distance. Neither the quartile distances nor the whiskers are sig-nificantly wide in most cases.

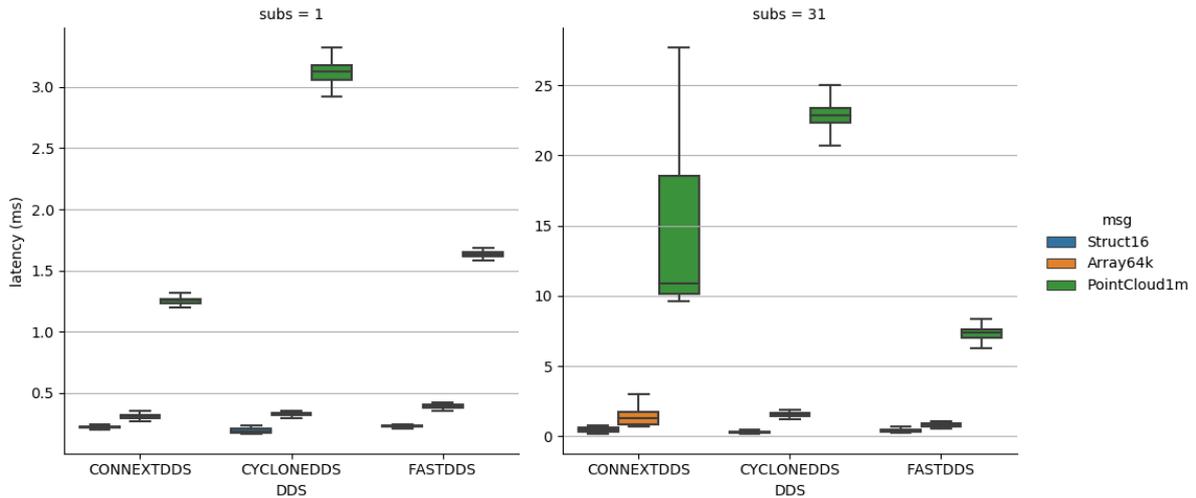

Figure 9: Comparison of the variance of the latency of the different DDS.

After an initial comparative measurement, CycloneDDS is evaluated again using tracing and an enlarged parameter space as an example, as CycloneDDS appears to be the most reliable, particularly in terms of low packet loss and low latency variance. Figure 10 again shows the performance of CycloneDDS for 1-N communication. The left-hand plot shows the measurement data for a frequency of 10Hz, while the right-hand plot shows the same measurements with a frequency of 100Hz. The x-axis shows the subscriber distribution (1, 7, 31, and 63), which is further divided according to packet size. For the frequency of 10 Hz, the latencies are still below the frequency limit on average, even if the whiskers exceed it, especially for the 2Mb packet. For the higher frequency of 100 Hz, the frequency limit of 10 ms is already exceeded for the 1 - 1 communication for the largest packet, and, as the number of subscribers increases, the 512kB and 1Mb packets also exceed this limit. It is also noticeable in this case that the quartile distance for these measurements is in most cases significantly larger than in the comparative measurements before. The increased frequency and the larger number of subscribers therefore show a strong influence on this.

It is also worth noting that the latency for 100 Hz, especially for 63 subscribers, shows a lower latency and lower variance. One reason for this behavior is due to the following correlation. Figure 11 shows the categorized latency of all received messages per number of subscribers and data size. The lower plot shows the view of the subscribers. As can be seen in the previous figure, almost every message arrives below the frequency limit of 100ms. These are categorized as "in time" in the plot. Only for the more complex configurations messages are occasionally lost or arrive too late.

However, the view of the publishers in the upper plot is conspicuous. Even for the simplest configuration, the publisher does not manage to send all messages in the given frequency time. This explains why the packet loss on the subscriber side remains so low despite the high load and large packet size. The majority of messages are not sent within the measurement window and therefore cannot be received on the subscriber side. As this behavior increases for higher frequencies, it is obvious that this is the reason for the better latencies in comparison. According to the frequency used and the measurement period, 600 messages should be sent in each configuration, but this is only possible for the two smallest packets in most cases. From 512kB upwards, the messages are increasingly delayed so that the total number of 600 is no longer reached in the measurement period.

Tracing can be used to determine where these delays occur. Figure 12 shows this broken down by the various layers of the ROS2 architecture on side of the publisher, again categorized by subscriber and data size. It is clear here that by far the most time is required at the DDS level, since the message is serialized and prepared for transmission at this level. It is already established that the serialization process takes a significant amount of time (Wang et al., 2018), especially as the message format of ROS2 and DDS is not uniform and therefore each requires its own processing. The figure also shows that the effect increases primarily with the data size, which also points to the serialization step.

The comparison between topologies 2 and 3 is also relevant: on the one hand, a publisher serves a larger number of subscribers, and, conversely, a large number of publishers serve a single node with many subscribers. This evaluates the efficiency of the executor in particular, in this case the SingleThreadedExecutor. Figure 13 shows this comparison: The left-hand side shows the latencies for 1-N communication, while the right-hand side shows the N-1 scenario. Both scenarios are almost identical, especially for 7 and 31 subscribers. For 63 subscribers, however, the difference is noticeably greater. Though, the influence of the ratio of publishers and subscribers on latency does not have a significant impact in general. The significantly greater variance for the N-1 scenario is due to the executor, as it processes all callbacks sequentially and therefore cannot process the open events quickly enough, especially under high load.

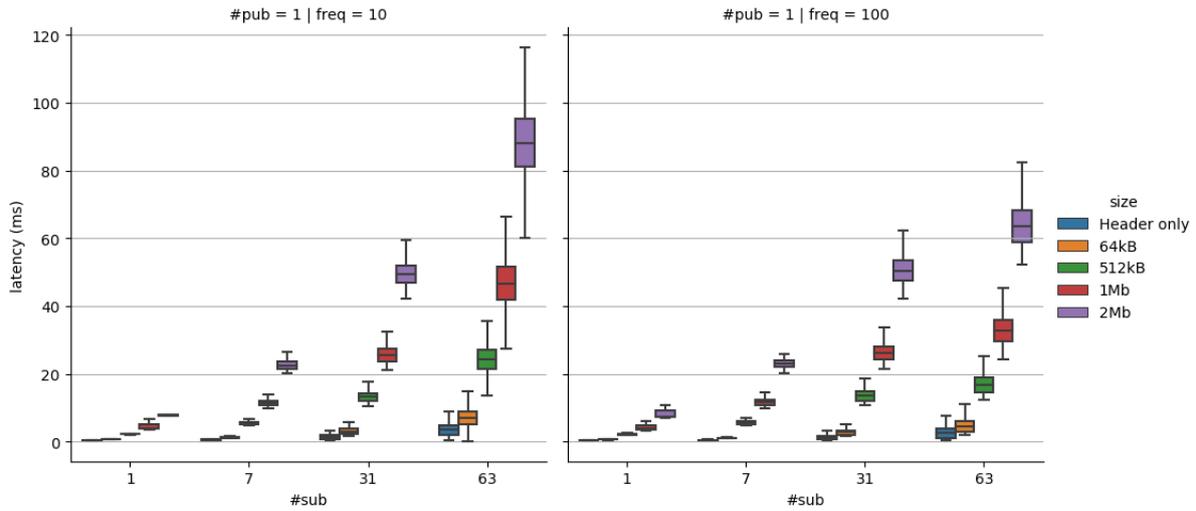

Figure 10: Latency per message size and number of subscribers.

Finally, it is checked whether the subscriber fairness described in (Maruyama et al., 2016), which is one essential change compared to ROS1, can also withstand more complex scenarios. Figure 14 shows this case as an example for a 1-N scenario, a packet size of 64kB, and a frequency of 10 Hz. Even if the latency for each subscriber differs slightly, they are on average max. 2 ms apart. There is also no staircase like increase for the subscribers, all have a latency of around 7 ms. The variance for this scenario is greater than for smaller scenarios, but here too none of the subscribers are significantly further apart than the others. This plot is comparable for all other combinations of parameters evaluated. This leads to the assumption that other influencing factors, such as frequency and size, have a stronger negative influence earlier, and therefore the performance collapses before the subscriber fair behavior can no longer be maintained.

## 6 SUMMARY

As the development of automated vehicles is an ongoing research task, this paper presents an evaluation of a ROS2 based ADS. The automated Mercedes E-Class of Fraunhofer FOKUS comprises both hardware and software components. The hardware setup consists of the sensor installation, the on-board PU for processing and planning, and the actuation hardware to control the vehicle. With a sensor rig, several cameras and LIDAR sensors are mounted on the roof of the vehicle. For vehicle control, a Drive-by-Wire system by Schaeffler Paravan is installed. The software components of the architecture are split in three segments: sensing, planning, and acting.

The complexity of the distributed nature of the ADS leads to the research question, if ROS2 fulfills the performance requirements for automated driving. Thus, a thorough analysis of ROS2 is performed for this paper. Two important aspects to consider are the latency, which measures the elapsed time between sending and receiving a message, and the packet loss, which measures the percentage of lost messages. The data size per message, number of messages sent per time unit, number of nodes, number of topics, and number of publishers and subscribers per topic are parameters of interest for the measurements.

Different DDS implementations are compared using a performance test framework, and one selected DDS is further evaluated using tracing to identify performance losses. The subscribers and publishers are started accordingly for of the three official DDS systems, FastDDS, CycloneDDS, and RTI Connext. Only the layers below the rmw layer are inspected for the comparative benchmark to isolate the influence of the DDS implementation. In the detailed benchmark tracing is used to track the message progress through the complete stack and to understand latency and message losses.

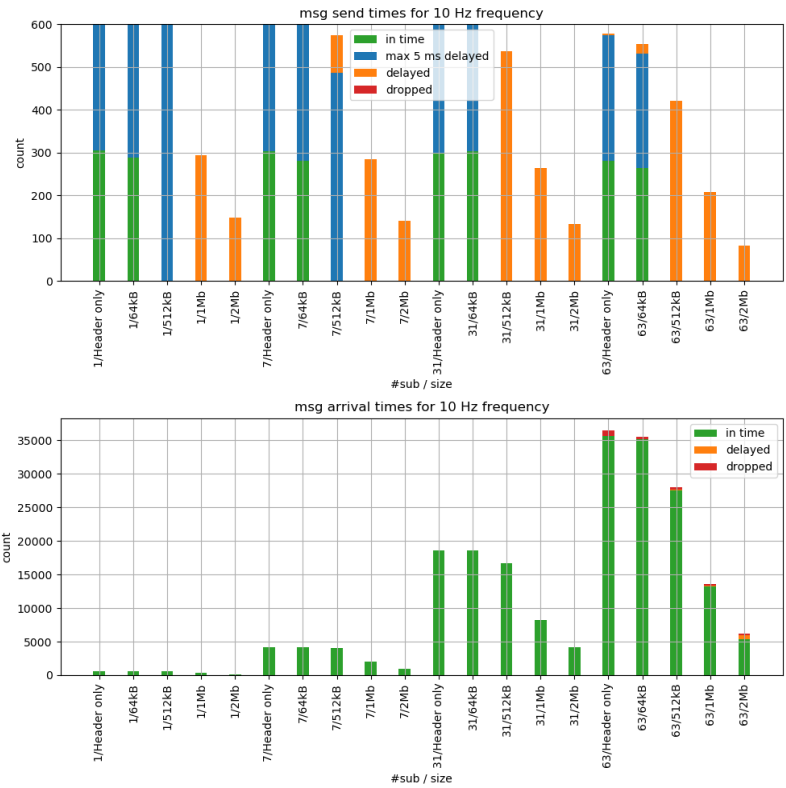

Figure 11: Overview of the messages sent and the associated arrival times.

Three different publish/subscriber topologies are assessed. The first one has a 1-1 relation between publishers and subscribers and topics, respectively. The next one has a 1-N publisher-subscriber relation, while the last one reverses this relation. They show comparable results for latency, error rate, and bandwidth. Latency depends mainly on packet size and number of nodes in the system. With high load, fragmentation of messages can lead to a lower packet loss. In general, packet loss is very low in the tested configurations.

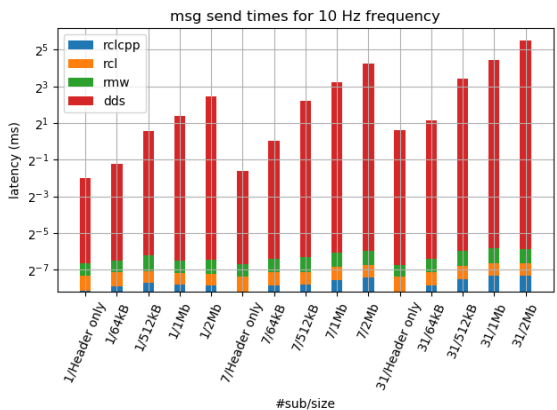

Figure 12: Overview of the messages sent and the associated arrival times.

A large part of the latency is generated on the publisher side before the actual sending and does not count into the transmission time. However, this affects the performance of the system, especially for high frequencies and large packet sizes. Latency remains very similar, comparing different topologies. Only in the n:1 scenario, the average latency is not changing much, while the variance increases significantly, due to the single threaded execution. In the 1:n scenario it can be observed that subscribers are served in a fair manner, and all have similar latency results. As overall both latency and packet loss are low in all tested setups, ROS2 proves as an efficient and reliable communication framework for an ADS. It should of course be noted that ROS2 does not support hard real time rigor. However, for the majority of communication, where low latency but no strict real-time capability is mandatory, it is a flexible communication framework that can be used to connect the components within an ADS. However, hard real time

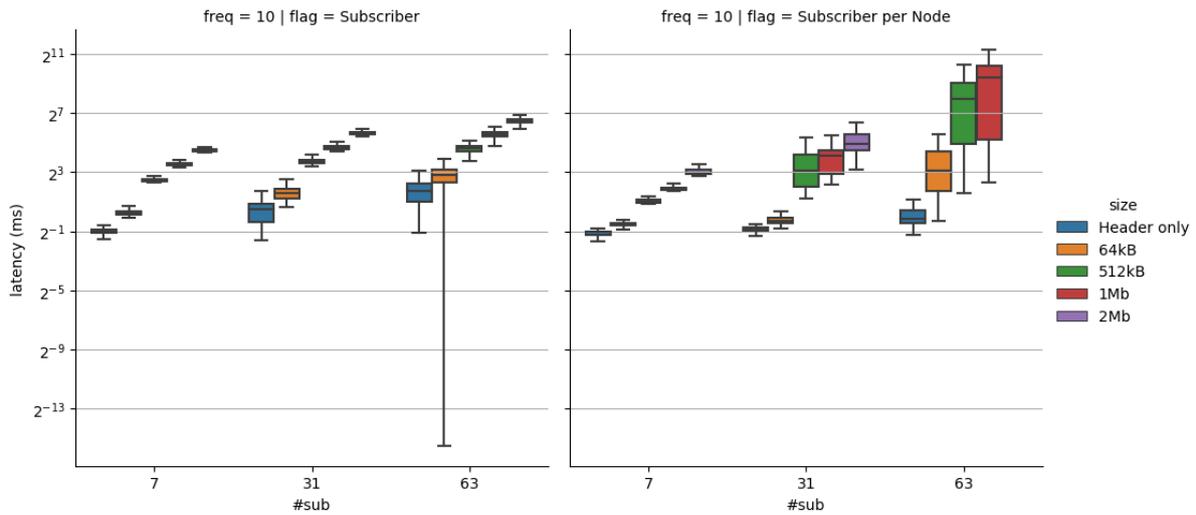

Figure 13: Comparison of latency for subscribers per node.

rigor should be implemented at least for the actuators.

Many other interesting measurements could be considered due to the various customization options available for ROS2 and the underlying DDS. This includes the influence of Quality of Service (QoS) profiles and their reliability with the coverage of various bandwidths, which is relevant for automated driving systems. Additionally, there are different execution models from single thread to multi threaded. This may require higher hardware requirements but can lead to significant performance improvements.

Another leverage point for performance is the specific configuration of the DDS used. All three implementations offer extensive options to adapt behavior for the scenario. For example, using Shared Memory (SHMEM) instead of UDP can avoid fragmentation of large messages and can reduce overall load. The Towards Zero Copy (TZC) technique presented in a study (Wang et al., 2018) eventually eliminates the overhead of serializing and copying messages.

## ACKNOWLEDGEMENTS

The work presented in this paper was conducted in the KIS'M project, funded by the German Federal Ministry for Digital and Transport (BMDV).

## REFERENCES

Baidu Apollo consortium (2023). Apollo Auto, an open autonomous driving platform. https://github.com/ApolloAuto/apollo/. Accessed: Dec. 01, 2023.

Bédard, C., Lajoie, P.-Y., Beltrame, G., and Dagenais, M. (2023). Message flow analysis with complex causal links for distributed ROS 2 systems. *Robotics and Autonomous Systems*, 161:104361.

Binder, T., Wedel, A., Bühren, M., Herget, C., Studer, S., Maier, H., Breu, J., Hafner, M., Hug, T., Hämmerling, C., et al. (2016). Assistenzsysteme in neuer Dimension. *Sonderprojekte ATZ/MTZ*, 21(Suppl 1):70–81.

Bédard, C., Lutkebohle, I., and Dagenais, M. (2022). ros2_tracing: Multipurpose low-overhead framework for real-time tracing of ROS 2. *IEEE Robotics and Automation Letters*, 7:6511–6518.

During, M. and Lemmer, K. (2016). Cooperative maneuver planning for cooperative driving. *IEEE Intelligent Transportation Systems Magazine*, 8(3):8–22.

Eiermann, L., Sawade, O., Bunk, S., Breuel, G., and Radusch, I. (2020). Cooperative automated lane

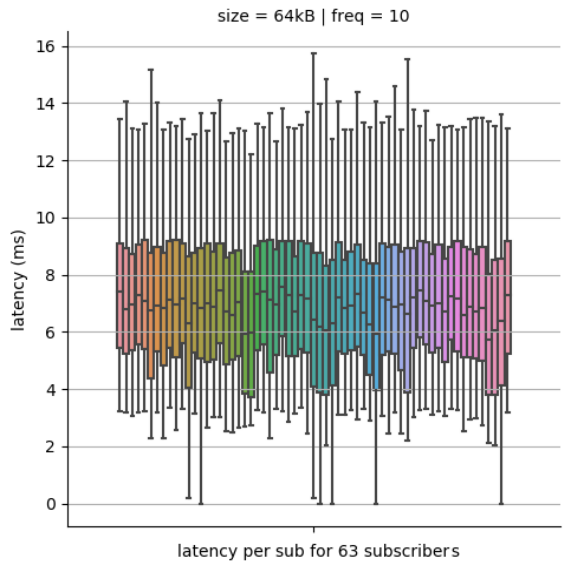

Figure 14: Comparison of latency for subscribers per node.

merge with role-based negotiation. In *2020 IEEE Intelligent Vehicles Symposium (IV)*, pages 495–501.

Granados, S. (2017). Who is chopping my application data and why should I care? https://www.rti.com/blog/2017/07/27/who-is-chopping-my-application-data-and-why-should-i-care/. Accessed: Dec. 04, 2023.

Gutiérrez, C. S. V., Juan, L. U. S., Ugarte, I. Z., and Vilches, V. M. (2018). Towards a distributed and real-time framework for robots: Evaluation of ROS 2.0 communications for real-time robotic applications. *arXiv preprint arXiv:1809.02595*.

Kronauer, T., Pohlmann, J., Matthe, M., Smejkal, T., and Fettweis, G. (2021). Latency analysis of ROS2 multi-node systems. *IEEE International Conference on Multisensor Fusion and Integration for Intelligent Systems*.

Lang, A. H., Vora, S., Caesar, H., Zhou, L., Yang, J., and Beijbom, O. (2019). Pointpillars: Fast encoders for object detection from point clouds. In *Proceedings of the IEEE/CVF conference on computer vision and pattern recognition*, pages 12697–12705.

Li, Z., Hasegawa, A., and Azumi, T. (2022). Autoware_perf: A tracing and performance analysis framework for ROS 2 applications. *Journal of Systems Architecture*, 123:102341.

Macenski, S., Foote, T., Gerkey, B., Lalancette, C., and Woodall, W. (2022). Robot operating system 2: Design, architecture, and uses in the wild. *Science Robotics*, 7.

Maruyama, Y., Kato, S., and Azumi, T. (2016). Exploring the performance of ROS2. In *Proceedings of the 13th International Conference on Embedded Software*, pages 1–10.

Object Management Group (2015). OMG Data Distribution Service (DDS) Version 1.4. https://www.omg.org/spec/DDS/1.4/PDF. Accessed: Dec. 04, 2023.

Open Robotics (2022a). About internal ROS 2 interfaces — ros 2 documentation: Rolling documentation. https://docs.ros.org/en/rolling/Concepts/About-Internal-Interfaces.html. Accessed: Dec. 11, 2023.

Open Robotics (2022b). Creating a workspace — ROS 2 documentation: Rolling documentation. https://docs.ros.org/en/rolling/Tutorials/Beginner-Client-Libraries/Creating-A-Workspace/Creating-A-Workspace.html. Accessed: Nov. 27, 2023.

Open Robotics (2022c). Using colcon to build packages — ROS 2 documentation: Rolling documentation. https://docs.ros.org/en/rolling/Tutorials/Beginner-Client-Libraries/Colcon-Tutorial.html. Accessed: Dec. 04, 2023.

Pemmaiah, A., Pangercic, D., Aggarwal, D., Neumann, K., and Marcey, K. (2022). Performance testing in ROS 2. https://www.apex.ai/post/performance-testing-in-ros-2. Accessed: Nov. 24, 2023.

Reif, K. (2011). *Bosch Autoelektrik und Autoelektronik*. Springer.

Reke, M., Peter, D., Schulte-Tigges, J., Schiffer, S., Ferrein, A., Walter, T., and Matheis, D. (2020). A self-driving car architecture in ROS2. *2020 International SAUPEC/RobMech/PRASA Conference, SAUPEC/RobMech/PRASA 2020*.

SAE (2021). J3016b: Taxonomy and definitions for terms related to driving automation systems for on-road motor vehicles—SAE International.

Samuel, M., Hussein, M., and Mohamad, M. B. (2016). A review of some pure-pursuit based path tracking techniques for control of autonomous vehicle. *International Journal of Computer Applications*, 135(1):35–38.

Schaeufele, B., Sawade, O., Pfahl, D., Massow, K., Bunk, S., Henke, B., and Radusch, I. (2017). Forward-looking automated cooperative longitudinal control: Extending cooperative adaptive cruise control (CACC) with column-wide reach and automated network quality assessment. In *2017 IEEE 20th International Conference on Intelligent Transportation Systems (ITSC)*, pages 1–6.

Schäufele, B., Sawade, O., Becker, D., and Radusch, I. (2017). A transmission protocol for fully automated valet parking using DSRC. In *2017 14th IEEE Annual Consumer Communications & Networking Conference (CCNC)*, pages 636–637.

Spielberg, N. A., Brown, M., Kapania, N. R., Kegelman, J. C., and Gerdes, J. C. (2019). Neural network vehicle models for high-performance automated driving. *Science robotics*, 4(28):eaaw1975.

The Autoware Foundation (2023). Autoware.auto. https://www.autoware.auto/. Accessed: Nov. 29, 2023.

Unseld, R. (2020). The next generation of vehicles will no longer have mechanical steering. *ATZelectronics worldwide*, 15(9):14–17.

Wang, Y.-P., Tan, W., Hu, X.-Q., Manocha, D., and Hu, S.-M. (2018). Tzc: Efficient inter-process communication for robotics middleware with partial serialization. *IEEE International Conference on Intelligent Robots and Systems*, pages 7805–7812.

Wulff, F., Schäufele, B., Sawade, O., Becker, D., Henke, B., and Radusch, I. (2018). Early fusion of camera and lidar for robust road detection based on U-Net FCN. In *2018 IEEE Intelligent Vehicles Symposium (IV)*, pages 1426–1431.

Yan, Y., Mao, Y., and Li, B. (2018). SECOND: Sparsely embedded convolutional detection. *Sensors*, 18(10):3337.